\documentclass[letterpaper, 10 pt, conference]{ieeeconf}
\IEEEoverridecommandlockouts
\usepackage[T1]{fontenc}
\usepackage{amsmath,amsfonts,amssymb}
\usepackage{array}
\usepackage{todonotes}
\usepackage{textcomp}
\usepackage{stfloats}
\usepackage{url}
\usepackage{graphicx}
\usepackage{booktabs}
\usepackage{cite}
\usepackage[bookmarks=true]{hyperref}
\usepackage{xcolor}
\usepackage{colortbl}

\definecolor{oursbg}{RGB}{214,230,248}
\definecolor{secbg}{RGB}{238,238,238}

\makeatletter
\def\bstctlcite{\@ifnextchar[{\@bstctlcite}{\@bstctlcite[@auxout]}}
\def\@bstctlcite[#1]#2{\@bsphack\@for\@citeb:=#2\do{\edef\@citeb{\expandafter\@firstofone\@citeb}\if@filesw\immediate\write\csname #1\endcsname{\string\citation{\@citeb}}\fi}\@esphack}
\makeatother
\hypersetup{pdfauthor={}, pdfcreator={}, pdfproducer={}}
\begin{document}
\bstctlcite{IEEEexample:BSTcontrol}

\title{\LARGE \bf
Real2Sem2Real: 3D Semantic Augmentations for Robot Learning from a Single Human RGB Demonstration
}

\author{Matthew Strong$^{1\ast}$, Ankush K. Dhawan$^{2\ast}$, Julien Poffet$^{3}$, Giuse Pham$^{2}$, Jinho So$^{2}$, and Monroe Kennedy III$^{2}$%
\thanks{This work has been submitted to the IEEE for possible publication. Copyright may be transferred without notice, after which this version may no longer be accessible.}%
\thanks{$^{1}$Department of Computer Science and $^{2}$Department of Mechanical Engineering, Stanford University, Stanford, CA, USA. $^{3}$D-MAVT, ETH Zurich, Switzerland; work done while visiting Stanford University. Emails: \{mastro1, ankushd, jpoffet, gpham123, jinhoso, monroek\}@stanford.edu.}%
\thanks{This research was supported by NSF Graduate Research Fellowship No. DGE-2146755.}%
\thanks{$^{\ast}$Both authors contributed equally to this work.}}

\maketitle
\thispagestyle{empty}
\pagestyle{empty}

\begin{abstract}
    Human video scales easily and does not need robot hardware, but existing methods require depth sensing, overlay the robot onto 2D video, or generate data only for the demonstrated objects. We present Real2Sem2Real (R2Sem2R), a real-to-semantic-to-real pipeline that turns one monocular RGB video of a human demonstration into thousands of robot demonstrations across the semantic object instances of a category. R2Sem2R tracks the SE(3) pose of each object with DINOv3-distilled Gaussian splats, outperforming existing methods and FoundationPose with estimated metric depth. R2Sem2R then aligns digital cousins, each generated from one or a few photographs, to the demonstrated objects with a semantic alignment. This alignment matches whole objects and the functional regions of target objects, such as a coffee machine's tray. R2Sem2R maps the human hand to a feasible robot grasp that extends to the cousins. Robot demonstrations are generated either in IsaacSim, for photorealistic rendering, or MuJoCo, for contact physics, and R2Sem2R can generate over 2,000 episodes in about two hours. On two real robots and four tasks, we train a vision-language-action policy on data from one human video per task, and policies trained on our data match or exceed  policies trained on teleoperation across diverse object instances. The project page is available at \url{https://real2sem2real.github.io/}.
\end{abstract}

\section{Introduction}

End-to-end robot learning maps pixels to actions in closed loop and is a promising route for robots that take on everyday tasks~\cite{kim2024openvla}, but it is data hungry. Vision-language-action (VLA) models are pretrained on more than a million robot trajectories~\cite{openx}, and Diffusion Policy~\cite{diffusionpolicy} still needs on the order of a hundred demonstrations per task. Robot data has two costs -- every trajectory requires human time, and every new object instance requires data collection on that instance, so hundreds of demonstrations of a single object will not generalize to other objects of the same type.

\begin{figure}[t]\centering
  \includegraphics[width=0.855\columnwidth]{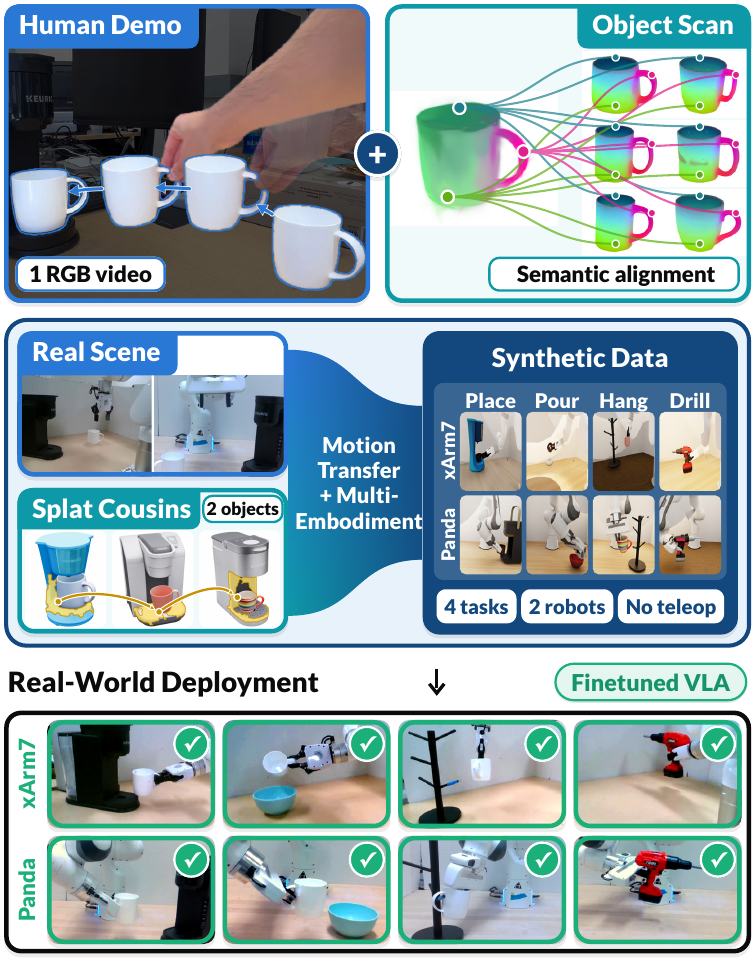}
  \caption{\textbf{Real2Sem2Real at a glance.} One human RGB video and a phone scan are the only inputs. DINOv3 part matches align demonstrated objects to digital cousins, including second objects such as coffee machines, and transferred motion generates data for many embodiments. A VLA fine-tuned on it runs on a real xArm7 and Franka Panda, comparable to teleoperation.}
  \label{fig:teaser}
\end{figure}
\begin{figure}[t]\centering
    \includegraphics[width=0.9\columnwidth]{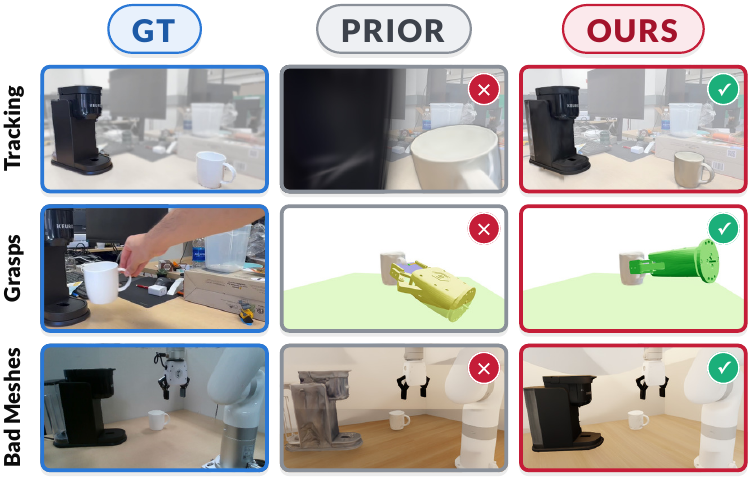}
    \caption{\textbf{Failure modes.} Prior trackers lose objects, prior grasps collide with the ground, and prior meshes have poor textures. R2Sem2R tracks the mug, recovers collision-free grasps, and renders realistic objects (GT: ground truth/real scene).}
    \label{fig:failure_modes}
  \end{figure}

Teleoperation~\cite{zhao2023act} is physically grounded but difficult to scale, and each alternative does not address every flaw. Demonstration augmentation~\cite{mimicgen, demogen} still starts from some teleoperation and needs privileged state or accurate 3D observations. Simulated collection~\cite{robocasa} needs modeled assets and a human in the loop, and handheld grippers~\cite{chi2024umi} require custom hardware and hundreds of demonstrations per task. Human video~\cite{wang2023mimicplay, kareer2024egomimic, qiu2025hat} must first be annotated with states and actions, which requires motion-capture and depth hardware~\cite{wang2024dexcap} or for rendering the robot into inpainted 2D frames~\cite{phantom}, where objects cannot be moved from their motion in video. Imprecise recovered object motion compounds into drifting poses and implausible geometry (Fig.~\ref{fig:failure_modes}) that are insufficient for a source of data.

Real-to-sim-to-real pipelines reconstruct real objects, recover their motion from a video, and render robot demonstrations in simulation with no robot in the loop. Real2Render2Real (R2R2R)~\cite{r2r2r} showed that one smartphone scan and one RGB video suffice to fine-tune a VLA, but it augments spatially, not semantically. For example, their exact mug and coffee machine in a human demonstration must be present at deployment, and covering more objects means a scan and a demonstration for every pairing. Digital cousins~\cite{acdc, simfoundry} substitute similar simulated assets, but overall similarity does not say where a mug should rest on a new coffee machine. Humans know a mug is lifted by its handle and set on a platform. \textbf{Given one human video of a mug placed on a coffee machine, why can't a policy succeed on mugs and coffee machines it has never seen?}

The first step is recovering object trajectories. The strongest pose estimators assume a depth sensor~\cite{foundationpose}, and RGB-only trackers drift or fail at initialization on textureless objects. We track each object with a DINOv3-distilled Gaussian splat~\cite{dinov3, 3dgs}, optimizing its pose against feature, photometric, silhouette, and monocular depth-ranking losses from an initialization seeded on the table plane. The grasp comes from the demonstrator's hand via HaMeR~\cite{hamer}, following the method in Phantom~\cite{phantom}.

The second step is augmentation. Beyond perturbing object poses, we add \textbf{semantic and functional augmentation}. Cousins are generated with MV-SAM3D~\cite{mvsam3d} from one or several images (online, AI-generated, or real photos), and distilled with DINOv3 features. A semantic alignment registers each cousin to the demonstrated object, and for functional placement/motion, a weighted alignment registers each functional and target object by its region of interaction rather than its whole body, so augmentation extends beyond single-object pick-and-place.

Finally, an interactive viewer can be employed to build the simulated scene, a camera optimization matches the simulated camera to the real one without AprilTags, and parallel generation in photorealistic IsaacSim~\cite{isaaclab}, or MuJoCo~\cite{mujoco}, generates over 2{,}000 episodes in about two hours, after about 25 minutes of setup, with no human in the loop. Among prior methods (Table~\ref{tab:taxonomy}), ours is the only one that transfers demonstrations across object instances from a \textit{single monocular RGB video.}

\begin{table}[t]
  \centering
  \caption{Demonstration generation methods. \emph{RGB only}: no depth or privileged state. \emph{Two-obj.}: carries a demonstrated two-object relation, such as a mug resting on a machine, to new instances of both. \emph{Cross-obj.}: transfer across a category. \emph{Demos}: source demos per task. \emph{Any arm}: any parallel-jaw arm.}
  \label{tab:taxonomy}
  \footnotesize\setlength{\tabcolsep}{2pt}\renewcommand{\arraystretch}{0.88}
  \newcommand{\hdr}[1]{\begin{tabular}[b]{@{}c@{}}#1\end{tabular}}
  \begin{tabular*}{\columnwidth}{@{}l@{\extracolsep{\fill}}cccccc@{}}
    \toprule
    Method & Demos & \hdr{No\\teleop} & \hdr{RGB\\only} & \hdr{Two-\\obj.} & \hdr{Cross-\\obj.} & \hdr{Any\\arm} \\
    \midrule
    MimicGen~\cite{mimicgen}   & 10 & $\times$   & $\times$   & $\times$   & $\times$   & \checkmark \\
    DemoGen~\cite{demogen}     & 1 & $\times$   & $\times$   & $\times$   & $\times$   & $\times$   \\
    Phantom~\cite{phantom}     & 250+ & \checkmark & $\times$   & $\times$   & $\times$   & \checkmark \\
    WARPED~\cite{warped}       & 30 & \checkmark & \checkmark & $\times$   & $\times$   & \checkmark \\
    R2R2R~\cite{r2r2r}         & 1 & \checkmark & \checkmark & $\times$   & $\times$   & \checkmark \\
    SimFoundry~\cite{simfoundry} & 10--15 & $\times$ & \checkmark & $\times$   & \checkmark & \checkmark \\
    \textbf{Ours}              & \textbf{1} & \checkmark & \checkmark & \checkmark & \checkmark & \checkmark \\
    \bottomrule
  \end{tabular*}
\end{table}

We present \textbf{Real2Sem2Real (R2Sem2R)}, summarized in Fig.~\ref{fig:teaser}. %
Our contributions are:

\begin{enumerate}
      \renewcommand{\labelenumi}{\textbf{\arabic{enumi}.}}
      \item An RGB-only 4D object tracker on DINO-distilled Gaussian splats that outperforms prior methods and FoundationPose with estimated metric depth;
      \item A semantic alignment of digital cousins for whole objects and functional placement/motion regions, producing demonstrations for every cousin pairing;
      \item Fast, parallel generation in MuJoCo or IsaacSim, enabled by an interactive scene creation  tool that ports directly to simulation; and
      \item Nearly \textbf{2{,}000} real rollouts on two robots, where a VLA trained on data from one human video per task matches or outperforms two hours of teleoperation.
\end{enumerate}

\section{Related Work}
\label{sec:related-work}

\textbf{Learning from human video.} The embodiment gap must be closed before human video can supervise a policy. DexCap~\cite{wang2024dexcap} uses motion-capture gloves and a depth camera, egocentric methods co-train with robot data~\cite{wang2023mimicplay, kareer2024egomimic, qiu2025hat}, and Phantom~\cite{phantom} overlays a rendered robot on the 2D video, tying each demonstration to its exact viewpoint and scene. VideoManip~\cite{videomanip}, VidAct~\cite{vidact}, and WARPED~\cite{warped} need less hardware but generate data only for the demonstrated objects in question, lacking generalization. WARPED builds a digital twin of the single object demonstrated per task and augments each demonstration tenfold, retexturing and translating the object, scaling the scene, and perturbing the camera, \textbf{but never changes the object's shape or geometry}. Learning from human video also requires many demonstrations per task, 250 to 350 videos for Phantom and 30 for WARPED (Table~\ref{tab:taxonomy}), whereas R2Sem2R starts from one.

\textbf{Object pose tracking.} BundleSDF~\cite{bundlesdf} and FoundationPose~\cite{foundationpose} track novel objects from RGB-D, so monocular pipelines substitute estimated depth, as VideoManip~\cite{videomanip} and VidAct~\cite{vidact} do with MoGe-2~\cite{moge2} and MoGe-3~\cite{moge3}, and accuracy hinges on that estimate (Sec.~\ref{sec:tracking-results}). RGB-only alternatives traditionally fit a differentiable renderer. RSRD~\cite{rsrd} renders distilled DINO features and fails on textureless objects in our experiments, and WARPED~\cite{warped} tracks a splat fitted to a SAM3D-generated mesh from a MegaPose~\cite{megapose} initialization, so its errors start from a generated model. We track a splat distilled from a scan of the real object, initialized upright on the table plane.

\textbf{Demonstration generation.} MimicGen~\cite{mimicgen}, RoboCasa~\cite{robocasa}, and DemoGen~\cite{demogen} multiply teleoperated demonstrations but need privileged state or depth, or some initial teleoperation. R2R2R~\cite{r2r2r} removes both by tracking objects in RGB video and replaying them kinematically, and its followers~\cite{r2rgen, real2edit2real, gaussianfactory, robosimgs} widen spatial augmentation. All vary \emph{where} the demonstrated objects are, not \emph{which}.

\textbf{Digital cousins and category-level generation.} ACDC~\cite{acdc} introduced digital cousins, and SimFoundry~\cite{simfoundry}, WorldComposer~\cite{worldcomposer}, and RoboCousin~\cite{robocousin} build sim-ready scenes and cousins from video, panoramas, or images, but take their trajectories from teleoperation, scripted skills, or planners, and SimFoundry registers cousins by overall geometry, which does not compute where a placed object must rest. Vision-language agents convert a recording into a simulatable twin~\cite{agenticreal2sim}, but each conversion often runs up to an hour per scene and reproduces only one scene in non-modular fashion.
However, some methods transfer a demonstration across \textit{instances}. Real2Gen~\cite{real2gen} uses two RGB-D cameras and places the object at the center of the target, which does not extend to a tray or a peg. ShapeGen~\cite{shapegen} collects five teleoperated RGB-D demonstrations per task, annotates contacts and keypoints, and trains a shape warping, changing only the \textit{manipulated object}. AffordGen~\cite{affordgen} transfers annotated keypoints with DINOv2 under the same assumptions. \textbf{R2Sem2R removes each requirement.} It starts from one monocular RGB human video with no teleoperation, annotation, or trained warping, and also varies the target object, in addition to the grasped object. Distilled feature fields~\cite{f3rm, rsrd} have helped with retrieval, segmentation, and grasp transfer, and we use DINOv3 features~\cite{dinov3}, distilled into each cousin's splat, to align whole objects and the contact region from the human demonstration autonomously (Sec.~\ref{sec:align}).

\section{Method}
\label{sec:method}
R2Sem2R takes one RGB video of a human performing the task, a phone scan of the demonstrated objects, and photographs of cousin objects. With one demonstration, object trajectories and valid grasps in the object's frame are computed, and we can use inverse kinematics (Sec.~\ref{sec:ik}) for any manipulator with a parallel-jaw gripper.

\subsection{Human Demonstration Parsing}
\label{sec:tracking}
\label{sec:ground}

\begin{figure*}[t]\centering
  \includegraphics[width=0.807\textwidth]{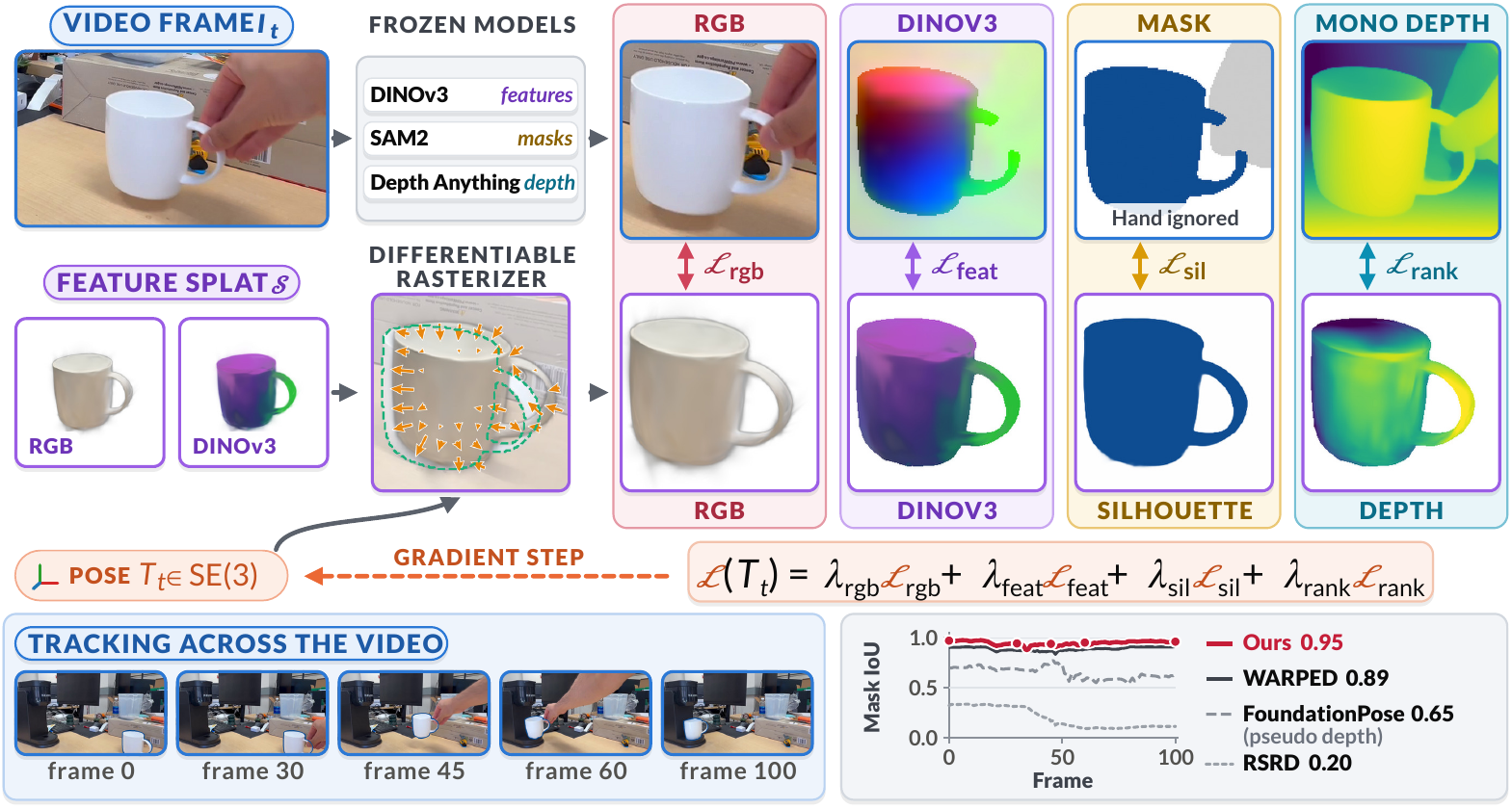}
  \caption{\textbf{Render-and-compare tracking.} Top: frozen DINOv3, SAM2, and Depth Anything turn a frame into RGB, features, a hand-free mask, and monocular depth; the splat is rendered at $T_t$ into the same channels and the losses update the pose. Bottom: the tracked mug and per-frame mask IoU for ours, WARPED, FoundationPose with predicted metric depth, and RSRD~\cite{rsrd}.}
  \label{fig:tracking}
\end{figure*}

Given an RGB video $V$, we want to track the $SE(3)$ pose of each object in every frame. Existing trackers need depth estimated from the frames, which is not robust (Sec.~\ref{sec:tracking-results}), so we track each object's DINOv3-distilled Gaussian splat $S$ by a render-and-compare method (Fig.~\ref{fig:tracking}).

\textbf{Object model.} Each object is scanned once with a phone and reconstructed as a Gaussian splat~\cite{3dgs} with light Pearson depth supervision. DINOv3 features~\cite{dinov3} are distilled onto every Gaussian from the training views~\cite{f3rm, rsrd}. To isolate the object, we segment it in the first scan image with a SAM3 text prompt~\cite{sam3} and propagate the mask $M_j$ to every view $j$. Gaussian $i$, with center $\mu_i$, receives the membership score $p_i = |\mathcal{V}_i|^{-1} \sum_{j \in \mathcal{V}_i} M_j(\pi_j(\mu_i))$, where $\pi_j$ projects into view $j$ and $\mathcal{V}_i$ is the set of views that see $\mu_i$. Splat $S$ keeps the Gaussians with $p_i > 0.5$ and $|\mathcal{V}_i| \geq 3$, which is robust to errors in any single mask. We find that this scoring gives us sharp, segmented splats of the demonstrated objects. We fit the table plane to the same scan with RANSAC, avoiding the tilt bias of monocular depth, and we scale $S$ to the measured height of the real object, so every tracked pose is metric.

\textbf{Tracking objective.} Let $T_t \in SE(3)$ be the object pose in the static camera frame at frame $t$. We render $S$ at $T_t$ and minimize $\mathcal{L}(T_t) = \sum_k \lambda_k \mathcal{L}_k$, $k \in \{\text{rgb}, \text{feat}, \text{sil}, \text{rank}\}$, by gradient descent (Fig.~\ref{fig:tracking}). Our loss consists of a \textit{mask-gated} DINOv3 feature loss for semantics, an RGB loss to refine it, a silhouette loss between rendered and detected masks, and a relative depth ranking loss using DepthAnything-V3~\cite{lin2025depth}.

\textbf{Initialization.} Because the scanned object rests on the table, we take advantage of this fact and seed the initialization of \textbf{upright orientations about the table normal}, and refine each seed with $\mathcal{L}$. We rank each seed after refinement by the one with the lowest combined RGB, DINOv3, silhouette (equal weighting), and highest IOU (double weighting).

\textbf{Hand occlusion.} During a grasp, the human hand often covers the asymmetric part of an object, like a mug handle. The silhouette loss pulls the pose to a wrong yaw, so we segment out the hand, and exclude its pixels from the loss. Later frames are also tracked from the prior pose with light smoothing. These tracked poses also map the table plane computed earlier into the camera frame for later stages.

\subsection{Hand-to-Gripper Retargeting}
\label{sec:grasp}

We compute valid parallel-jaw grasps from the human hand grasp with RGB only.

\textbf{Hand registration.} HaMeR~\cite{hamer} recovers the hand in every frame. The HAMER hand mesh is metric, but it sits roughly half a meter off the scene due to the monocular scale ambiguity. We thus fit an affine map from 
monocular depth~\cite{lin2025depth} to the metric splat's rendered depth on the object's pixels, compute the hand's depth from the transformation, and finally shift the mesh along its viewing rays. We then apply a small depth offset, which is tuned automatically with a 1-D sweep to minimize the fingertip-to-object distance, which improves the initial prior hand to object pose. 

\textbf{Gripper mapping.} Following Phantom~\cite{phantom}, the tool center point is the midpoint of the thumb and index fingertips, the closing axis is the line between them, and the approach is the direction from the wrist to that midpoint. The grasp $G \in SE(3)$ is stored in the object's frame, so the end-effector target at frame $t$ is $T_t\,G$.

\textbf{Best grasp.} Over the video the hand approaches, closes on, carries, and releases the object, so the grasp is taken from the carry. A frame qualifies if the thumb and index fingertips are closed in a pinch, the wrist is rigid in the object's frame, and the object is lifted more than 2\,cm above its rest height on the table. In each of these frames, we sample 100 $SE(3)$ perturbations of the retargeted gripper, within 2\,cm and $45^\circ$, and visit them in order of distance from the hand's pose. We can filter out collisions with the meshes of the gripper, TSDF mesh of the splat, and using the table plane. The first collision-free antipodal grasp found, with the gripper fully open, is the best grasp at that frame, and across frames we keep the one whose gripper and object-mesh normals are most nearly opposed. This grasp can then be augmented into additional grasps and to enforce grasp depth.

\subsection{Semantic Augmentation}
\label{sec:align}
\label{sec:cousins}

\begin{figure*}[t]\centering
  \includegraphics[width=0.79\textwidth]{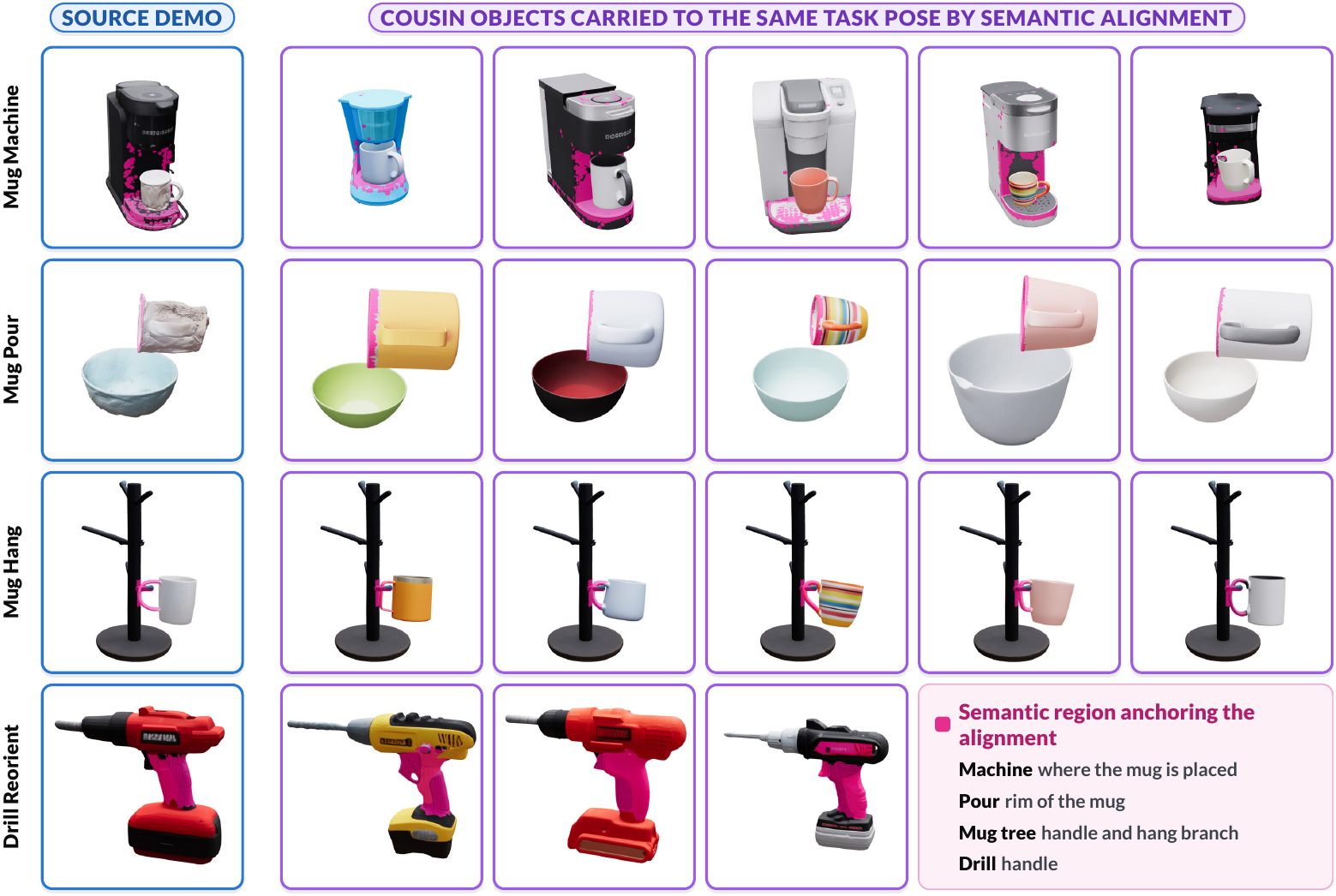}
  \caption{\textbf{Semantic alignment across tasks.} Source objects at the task pose (left, blue) and cousins carried there by semantic alignment (right). Pink marks the region anchoring the task-weighted alignment: the machine's tray, the poured mug's rim, the mug handle and its branch, and the drill's handle.}
  \label{fig:alignment}
\end{figure*}

The parsed demonstration covers one object instance per category. To cover others, each cousin must be functionally aligned to the demonstration. The alignment has three stages, each repairing a failure of the others (Fig.~\ref{fig:alignment}).

\textbf{Cousin assets.} Real2Sem2Real only needs one or a few photographs of cousins assets, which can be found in the real world, online stores, or AI generated with a text prompt. SAM3~\cite{sam3} segments the object and SAM3D~\cite{sam3d} or MV-SAM3D~\cite{mvsam3d} is used for several views. These methods generate a splat and a textured mesh, and the generation of ten cousins takes about seven minutes on one RTX~5090. Since the assets lack metric scale, we scale each cousin to its real height taken from the product specification, estimated by a VLM, or measured with the real-life object. We apply the same DINOv3 feature distillation to these objects.

\textbf{Semantic ICP.} We first align with bidirectional ICP (iterative closest point) from a few candidate orientations. We then \textit{cluster} the distilled features with K-means, and match clusters across objects with cosine similarity. Our \textit{semantic ICP}  restricts ICP correspondences to the matching clusters and uses a summed feature and spatial distance, so that semantically similar parts of objects pair together.

\textbf{Task-weighted ICP.} Whole-object alignment does not align the part that matters for the demonstration, such as a coffee machine tray with a mug on it. We use a VLM~\cite{claudeopus55} to select the key frame of the demonstration, which is the end state for a placement, or the start of a continuous motion, like a pour. The goal is that, given the target state of the demonstration, how can we vary cousins on both objects so that the semantic meaning is unchanged?

For a placement, the key region is the sphere of radius $r$ of both objects' Gaussians at the settled placement. Based on the task description, the VLM outputs if a settled alignment is needed, in which the grasped object is aligned positionally to match the bottom of the original grasped object at the target state. For the target object, and in cases where settled alignment (such as mug tree hang) is not the case, this region is used. For a motion such as a pour, the VLM and SAM3 segment the aligned part of the grasped object -- the rim of the mug -- and the same sphere test is applied against the target object. A DINOv3 cosine-similarity threshold grows the region to cover the semantic part.\textbf{ We transfer the region to each cousin by DINOv3 feature similarity} and realign with its points weighted $20\times$, which anchors the placement to the semantically relevant region (Fig.~\ref{fig:alignment}). Our alignment sometimes creates unwanted tilts to the object. We project the rotation onto a yaw $\psi$ about the table normal and re-translate so the weighted region stays fixed.

\textbf{Trajectory transfer.} With $A$, which is the rigid alignment of a cousin to its reference, the cousin inherits the tracked pose $T_t^{\text{cousin}} = T_t\,A$, and the goal of the manipulated object relative to the target object holds for every pairing of cousins. We additionally apply a ground alignment step, which moves the grasped object only in $z$ so that it starts level on the table; this does not change the semantic alignment used for the target states of the demonstration.

\subsection{Grasp Transfer and Inverse Kinematics}
\label{sec:ik}

\begin{figure*}[t]\centering
  \includegraphics[width=0.95\textwidth]{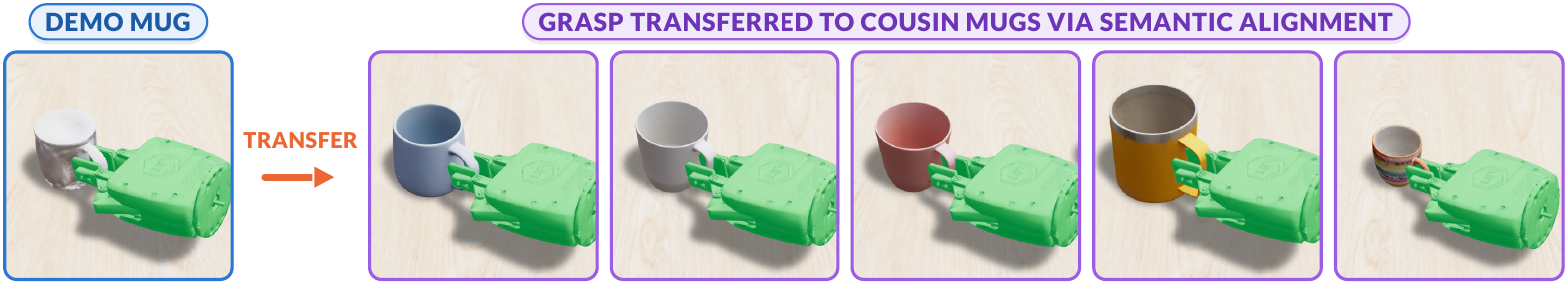}
  \caption{\textbf{Grasp transfer.} The grasp retargeted from the human hand on the scanned mug (left) is carried to each cousin by its semantic alignment (right), closing on the handle of every one.}
  \label{fig:grasptransfer}
\end{figure*}

\textbf{Grasp transfer.} The grasp $G$ of Sec.~\ref{sec:grasp} lives in the reference frame, so it follows each cousin through its alignment before the final ground alignment (Fig.~\ref{fig:grasptransfer}). We check each transferred grasp for force closure and collision on the cousin's mesh,  and we resample locally if it collides.

\textbf{Inverse kinematics.} For robot base pose $B$, the end-effector targets are $B^{-1}\,T_t\,G$, preceded by a collision-free pre-grasp offset along the approach axis. We solve with jaxmp and discard episodes that miss targets or self-collide. 

\subsection{Scene Construction and Camera Optimization}
\label{sec:scene}

We build a simulated scene matching the real setup and recover the camera pose from one RGB frame of the robot at home, with the robot as the fiducial, instead of AprilTag.

\begin{figure}[t]\centering
  \includegraphics[width=\columnwidth]{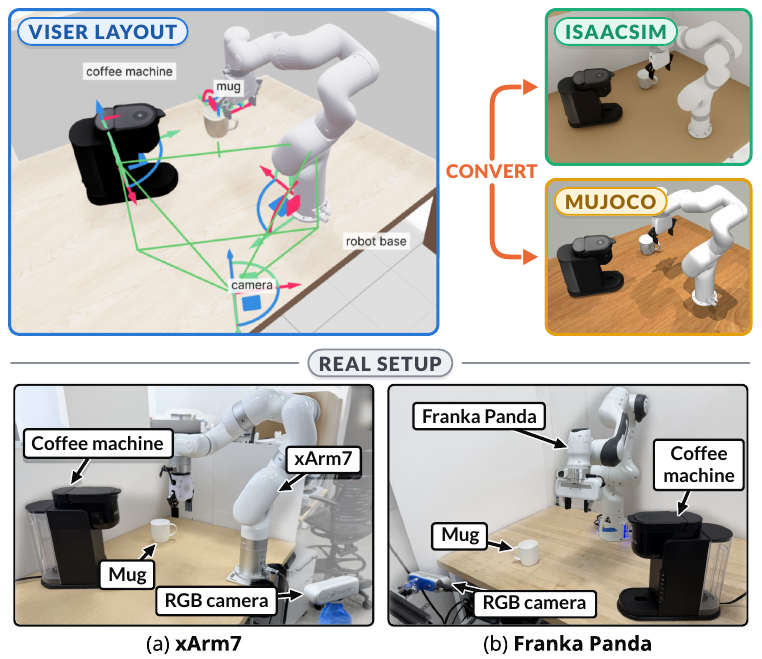}
  \caption{\textbf{Scene construction and real setup.} Top: the Viser layout is converted into IsaacSim (photorealistic rendering, kinematic carry) and MuJoCo (contact physics) scenes. Bottom: the real (a) xArm7 and (b) Franka Panda, each with one external RGB camera as the policy's only image input.}
  \label{fig:viser}
\end{figure}

\textbf{Scene.} A user places the robot, table, and posterboard in an interactive tool built on Viser \cite{yi2025viser}, with tape measures to match offsets. The table is the plane of Sec.~\ref{sec:ground}, objects are seated on it at metric scale, and the base is placed so the trajectory is within reach (Fig.~\ref{fig:viser}). The saved layout ports directly to IsaacSim and MuJoCo, so both simulators share one scene. Each object is a textured mesh with its alignment and metric scale baked in. MuJoCo additionally decomposes it into convex collision hulls within 3\,mm of the surface.

\textbf{Camera objective.} We automatically calibrate the camera pose of our scene, initialized from the user in the Viser tool, and optimized in IsaacSim \cite{isaaclab}. We optimize the camera extrinsics to minimize $\mathcal{L}_{\text{cam}}(\theta) = 1 - \mathrm{IoU}(M_\theta, M^\ast) + \lambda_{\text{rgb}} \mathcal{L}_{\text{rgb}} + \lambda_{\text{feat}} \mathcal{L}_{\text{feat}}$, where $M_\theta$ and $M^\ast$ are the SAM3 masks of the robot in the rendered and real images. The silhouette IoU dominates, and the masked RGB and DINOv3 terms give a signal when it stagnates. Because the simulator renderer is non-differentiable,  we estimate the gradient by simultaneous perturbation, two renders per step regardless of dimension, in coarse-to-fine passes about the best pose.

\subsection{Data Generation}
\label{sec:transfer}
\label{sec:replay}

\textbf{Episode sampling.} Each episode selects a pairing of cousins and new poses for both objects from uniform ranges on the table. We warp the demonstrated trajectory to connect the new start to the 
transferred goal $T_t\,A$, with the start offset decaying to zero along the arc length, so the motion at the goal is exactly as demonstrated.

\textbf{Simulation.} The two simulators trade visual realism against physical realism (Fig.~\ref{fig:viser}). IsaacSim~\cite{isaaclab} renders photorealistic images of many environments in parallel and carries the object kinematically with the gripper. MuJoCo~\cite{mujoco} simulates contact, from settling to the grasp and the release under gravity, and keeps an episode only if the object rests within tolerance of its goal, but renders less realistically. Each episode randomizes table texture, lighting, skybox, and the camera pose about the calibrated one; the real camera stays fixed during evaluation. Samples discarded for an IK failure, a collision, or a failed MuJoCo settle are counted in the generation time, which yields about 2{,}000 kept episodes in about two hours on one RTX 5090. We record the RGB frame, the end-effector pose from forward kinematics of the manipulator, and the commanded gripper opening at $10$\,Hz.

\section{Experiments}
\label{sec:experiments}

\subsection{Experimental Setup}
\label{sec:setup}

\begin{figure}[t]\centering
  \includegraphics[width=0.874\columnwidth]{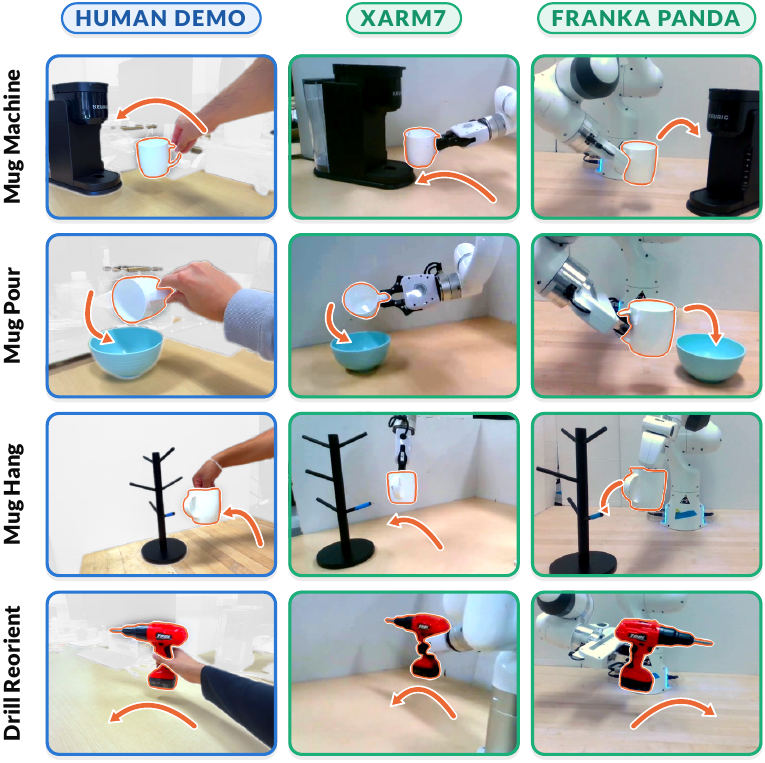}
  \caption{\textbf{Tasks.} A single RGB human demonstration (left) generates training data for both robots. Real rollouts conducted on the xArm7 and Franka Panda (right). The manipulated object is outlined in orange with its motion.}
  \label{fig:tasks}
\end{figure}

\textbf{Tasks.} We evaluate four tasks, each emphasizing a different part of the pipeline (Fig.~\ref{fig:tasks}). \emph{Place mug on coffee machine} is our headline task, and both the mug and the machine vary across cousins. \emph{Pour the mug into the bowl} is a continuous motion that requires a correct trajectory to avoid spilling, and the mug and bowl vary. \emph{Hang the mug on the mug tree} succeeds only if the handle clears the peg at the correct orientation, a tighter contact than a tray set-down; the mug varies and the tree is fixed. \emph{Drill reorient} picks up a drill and sets it upright at a target orientation, testing grasp recovery on objects beyond mugs (Sec.~\ref{sec:grasp}); the drill varies.

\textbf{Hardware.} All tasks run on a UFactory xArm7 and a Franka Emika Panda, each with its native gripper (Fig.~\ref{fig:viser}, bottom) (we 3D print a part for the Panda's fingers to improve surface coverage). An external Intel RealSense D435i, in which we only use RGB, is placed behind the arm for the xArm7 and in front of it for the Panda.

\textbf{Policy.}
\label{sec:training}
We fine-tune the base $\pi_{0.5}$~\cite{pi05} with LoRA. The policy takes a third-person $224\times224$ RGB frame and an $8$-D proprioceptive state, the end-effector position and axis-angle orientation in the robot base frame plus the gripper opening, and predicts a chunk of $10$ actions at $10$\,Hz. Each action is $7$-D, consisting of position change, axis-angle rotation from the current to the next orientation, and gripper command.
We train with AdamW at batch size 32 and a cosine schedule from $2.5\times10^{-5}$ to $2.5\times10^{-6}$ after 1{,}000 warmup steps, for 5 epochs on the generated datasets, 58k to 93k steps, since episode duration varies by task. We train to 30k steps on the teleoperated datasets, the default fine-tuning length of the $\pi_{0.5}$ release~\cite{pi05}, and we find that training additional steps does not yield additional increase in performance.

\textbf{Conditions.} On each task and robot, we compare \textbf{Teleop}, 150 teleoperated demonstrations across several instances of each category; \textbf{R2R2R+}, a strengthened R2R2R~\cite{r2r2r}, 2{,}000 generated episodes of the single demonstrated pair; and \textbf{R2Sem2R}, 2{,}000 generated episodes across cousin pairings, each cousin set being one reference object plus its SAM3D cousins. R2R2R+ keeps only the demonstrated objects, as R2R2R and WARPED~\cite{warped} do, but runs through our pipeline. Our tracker replaces theirs (we show stronger performance in ours with Sec.~\ref{sec:tracking-results}), our grasp selection is collision free, and the simulators and IK are the same. R2R2R+ can be considered as a strengthened baseline, or R2Sem2R without cousins. WARPED's retexturing changes appearance but, unlike a cousin, not geometry. In a data scaling experiment, we train on 50, 100, and 150 teleop demonstrations, and 50, 100, 150, 666, 1333, and 2000 episodes for R2Sem2R

\textbf{Test objects.} The coffee machine task is tested on seven mugs and seven coffee machines. Our teleoperation trains on four of each, and the other three of each are held out. R2Sem2R's cousins set add further mugs and machines (16 mugs and 9 coffee machines). The pour task uses the same split over seven mugs and seven bowls. Hang mug task is similar, but uses only one mug tree. Drill reorient uses four drills, all of which appear in the training data of both teleoperation and R2Sem2R, so it compares the two data sources on identical objects.

\textbf{Protocol.} For each task and robot, we run 50 real-robot trials (40 for drill). Objects are placed at random within the spatial distribution of the training data. A trial succeeds if, after release, the mug sits level on the coffee machine's tray, so that coffee would fall into it without spilling; the mug pours over the bowl, so that liquid would land in it; the mug hangs on a branch of the mug tree; or the drill stands upright within the target orientation.

\subsection{Object Tracking Accuracy}
\label{sec:tracking-results}
We compare tracking against the closest alternatives on the seven objects of our four tasks, by mean per-frame mask IoU against the same SAM masks. Every method receives the same object model, a TSDF mesh fused from our splat's renders at the measured object size. WARPED~\cite{warped} is run as specified, and RSRD~\cite{rsrd} is re-implemented in our tracker with its released settings. FoundationPose~\cite{foundationpose} is RGB-D by design, so we give it the best monocular metric depth of~\cite{spatrackerv2, pi3, lin2025depth, depthanythingv2} under two conditions: raw, which uses nothing of ours, and calibrated to our splat at our tracker's poses once per video, an upper bound. On our videos, our tracker is the most accurate on every object under every condition. Visualizations are in the supplemental material.

We also evaluate on YCBInEOAT~\cite{se3tracknet}, whose static camera matches our setting. In it, a robot moves five YCB objects through nine sequences. In this experiment, every method starts from the ground-truth first pose, FoundationPose uses MoGe-3~\cite{moge3} depth, and accuracy is the AUC (0--10\,cm) of ADD, the mean model-point error, and ADD-S is its symmetry-tolerant variant. On YCBInEOAT, our tracker leads every monocular baseline on average (Table~\ref{tab:tracking}). 

\begin{table}[t]
  \centering
  \caption{Tracking accuracy. Top: mask IoU on our objects. Bottom: AUC (\%) on YCBInEOAT, starting from the ground-truth pose, with the per-video depth scale set on the first frame. FoundationPose with depth gets 95.8 ADD-S and 93.5 ADD.}
  \label{tab:tracking}
  \footnotesize\setlength{\tabcolsep}{2.85pt}\renewcommand{\arraystretch}{0.9}
  \begin{tabular}{l>{\columncolor{oursbg}}ccccc}
    \toprule
    & \multicolumn{1}{c}{} & & & \multicolumn{2}{c}{FoundationPose~\cite{foundationpose}} \\
    \cmidrule(l){5-6}
    Object & \textbf{Ours} & WARPED & RSRD & per-video & raw \\
    \midrule
    \multicolumn{6}{>{\columncolor{secbg}}l}{\textit{Our videos: mask IoU}} \\
    Mug (to machine)        & \textbf{0.949} & 0.890 & 0.197 & 0.897 & 0.648 \\
    Coffee machine          & \textbf{0.916} & 0.894 & 0.861 & 0.872 & 0.680 \\
    Mug (to tree)           & \textbf{0.904} & 0.000 & 0.097 & 0.864 & 0.750 \\
    Tree stand              & \textbf{0.840} & 0.712 & 0.505 & 0.757 & 0.601 \\
    Mug (pour)              & \textbf{0.935} & 0.900 & 0.025 & 0.579 & 0.443 \\
    Bowl (pour)             & \textbf{0.982} & 0.849 & 0.503 & 0.972 & 0.692 \\
    Drill                   & \textbf{0.864} & 0.525 & 0.714 & 0.819 & 0.587 \\
    \midrule
    Mean                    & \textbf{0.913} & 0.681 & 0.415 & 0.823 & 0.629 \\
    \midrule
    \multicolumn{6}{>{\columncolor{secbg}}l}{\textit{YCBInEOAT~\cite{se3tracknet}: mean AUC over five objects}} \\
    ADD-S                   & \textbf{51.4} & 44.0 & 5.1 & 30.3 & 3.0 \\
    ADD                     & \textbf{41.3} & 25.1 & 1.9 & 22.8 & 1.5 \\
    \bottomrule
  \end{tabular}
\end{table}

\subsection{Real-Robot Results}
\label{sec:policy-learning}
\label{sec:task-results}

\begin{figure}[t]
  \centering
  \includegraphics[width=0.86\columnwidth]{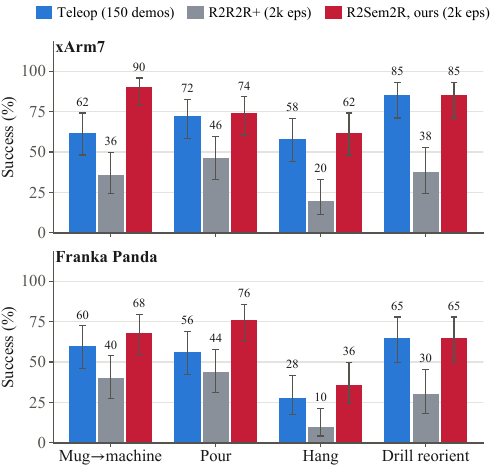}
  \caption{\textbf{Real-robot success rate} on the xArm7 (top) and Franka Panda (bottom). Whiskers: Wilson 95\% interval.}
  \label{fig:success}
\end{figure}

Fig.~\ref{fig:success} gives the success rate of every policy. R2Sem2R and R2R2R+ are generated from the same human video, 2{,}000 episodes each, and differ only in whether the episodes span cousins, so their gap isolates semantic augmentation. On every task and robot, R2Sem2R leads R2R2R+ by 26--54 points and matches or exceeds teleoperation.  Drill reorient isolates data quality from object coverage, since both policies train on the same four drills. 

\subsection{What Drives Performance}
\label{sec:analysis}
\begin{figure}[t]
  \centering
  \includegraphics[width=0.86\columnwidth]{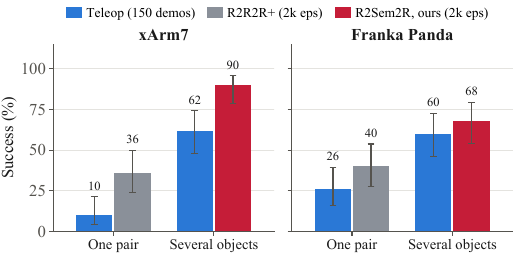}
  \caption{\textbf{Object diversity versus data source} on placement. \emph{One pair}: 150 teleop demos or 2{,}000 R2R2R+ episodes on demonstrated pair. \emph{Several objects}: teleop on 4x4 pairs, or R2Sem2R. Whiskers: Wilson 95\% interval.}
  \label{fig:diversity}
\end{figure}

\textbf{Object diversity versus data source.} On placement (Fig.~\ref{fig:diversity}), object diversity matters more than data source. Teleoperation and R2R2R+ on a single mug and machine are the two weakest policies on both robots, so spatial augmentation alone does not substitute for new objects.

\begin{figure}[t]
  \centering
  \includegraphics[width=0.86\columnwidth]{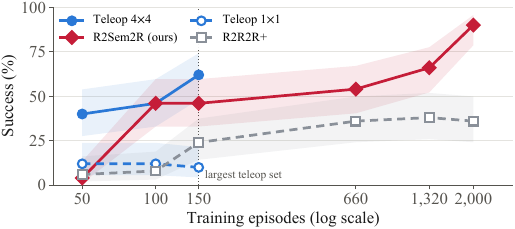}
  \caption{\textbf{Success versus training-set size} on xArm7 placement. Solid: several objects; dashed: demonstrated pair; bands: Wilson 95\% interval.}
  \label{fig:datascaling}
\end{figure}

\textbf{Data scaling.} On xArm7 coffee machine (Fig.~\ref{fig:datascaling}), both single-pair policies plateau (teleoperation at 10--12\%, R2R2R+ at 36--38\% from 660 episodes on), while R2Sem2R keeps improving. A generated episode is worth \textit{less} than a teleoperated one, but R2Sem2R matches the best teleoperated policy at 1{,}320 episodes and passes it at 2{,}000.

\textbf{Simulator.} Generating the same 2{,}000 episodes in MuJoCo instead of IsaacSim lowers xArm7 placement from 45/50 (90\%) to 31/50 (62\%), the same as teleoperation, so for this RGB policy, visual realism likely matters more than contact dynamics. Other generated-data results use IsaacSim.
\subsection{Collection Cost}
\label{sec:cost}
\begin{figure}[t]
  \centering
  \includegraphics[width=0.86\columnwidth]{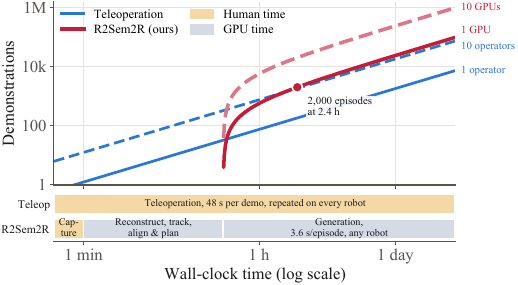}
  \caption{\textbf{Demonstrations against wall-clock time} (log-log) for 1 and 10 teleoperators or GPUs; R2R2R+ coincides with R2Sem2R and is omitted. Strips show where each method's time goes, shaded as human or GPU time. Robot setup, shared by all methods, is not shown.}
  \label{fig:scaling}
\end{figure}

Teleoperation grows linearly with human time at about 48\,s per demonstration, including scene resets. R2R2R+ and R2Sem2R instead need about 1 minute of human time to capture the video and scan, and about 25 minutes of GPU time (20 minutes for R2R2R+, which has no cousins) to reconstruct the objects, track the demonstration, generate and align the cousins, and plan the robot motion, before generating 2{,}000 episodes in about two hours on one GPU, about 3.6\,s per episode (Fig.~\ref{fig:scaling}). The 150 teleoperated demonstrations cost about two hours of operator time; the 2{,}000 R2Sem2R episodes cost about a minute of human time and 2.5 hours end to end. Both also need the same 30-minute robot setup, so it is left out. Cousin photographs add no capture time (Sec.~\ref{sec:cousins}). Policy training does vary for R2Sem2R, but with widely available compute, human RGB videos still scale better than teleoperated demos.

\section{Conclusion}
\label{sec:conclusion}
We presented R2Sem2R, which generates semantically augmented demonstrations from one human RGB video. One human video becomes robot-independent data across many cousins, augmenting both functional and target objects. Policies trained on R2Sem2R data perform robustly in the real world, matching or exceeding teleoperation.

\textbf{Limitations.} R2Sem2R requires a metric measurement of the demonstrated objects, and monocular depth proved unsuitable. IsaacSim's kinematic replay omits finger-object contact, which limits contact-rich tasks. Future work will combine MuJoCo dynamics with IsaacLab or Blender rendering. 
R2Sem2R handles rigid objects only. Recovering a screw axis from the demonstration~\cite{screwsplat} and transferring it through the alignment is future work for articulated cousins.

\bibliographystyle{IEEEtran}
\bibliography{references}

\end{document}